\documentclass[10pt,journal,compsoc]{IEEEtran}

\ifCLASSOPTIONcompsoc
  \usepackage[nocompress]{cite}
\else
  \usepackage{cite}
\fi
\ifCLASSINFOpdf
  \usepackage[pdftex]{graphicx}
\else
\fi
\usepackage{grffile}
\usepackage{amsmath,amsfonts}
\usepackage{amsthm}

\usepackage{algorithm}
\usepackage{algorithmic}
\usepackage{array}
\usepackage[colorlinks=true,linkcolor=blue,citecolor=blue]{hyperref}
\usepackage{xspace}
\usepackage{booktabs}
\usepackage{graphicx}
\usepackage{hyperref}

\ifCLASSOPTIONcompsoc
 \usepackage[caption=false,font=footnotesize,labelfont=sf,textfont=sf]{subfig}
\else
 \usepackage[caption=false,font=footnotesize]{subfig}
\fi

\usepackage{url}

\usepackage[capitalize,noabbrev]{cleveref}
\usepackage{xcolor}
\usepackage{xspace}
\usepackage{multirow}
\usepackage{colortbl}
\usepackage{wrapfig}
\usepackage{bm}
\usepackage{booktabs}

\mathchardef\mhyphen="2D

\usepackage{enumitem}

\usepackage{makecell}
\usepackage{tikz}
\usetikzlibrary{cd}

\usepackage[T1]{fontenc}%
\usepackage{xhfill}%

\usepackage{pifont}

\definecolor{Gray}{gray}{0.9}

\definecolor{Darkgray}{rgb}{0.2,0.2,0.2}
\definecolor{tabgreen}{rgb}{0,0.6,0}
\definecolor{tabred}{rgb}{0.8,0,0}
\definecolor{taborange}{rgb}{0.8,0.4,0}
\definecolor{tabpurple}{rgb}{0.5,0,0.5}
\definecolor{red}{rgb}{0.8,0,0}
\definecolor{brown}{rgb}{0.6,0.3,0}
\definecolor{blue}{rgb}{0,0,0.8}
\definecolor{magenta}{rgb}{0.8,0,0.8}
\begin{document}

\title{Pruning via Merging: Compressing LLMs via Manifold Alignment Based Layer Merging}

\author{Deyuan Liu$^{\ast}$, 
        Zhanyue Qin$^{\ast}$, 
        Hairu Wang, 
        Zhao Yang, 
        Zecheng Wang, 
        Fangying Rong,
        Qingbin Liu,\\
        Xi Chen,
        Cunhang Fan, 
        Zhao Lv,
        Zhiying Tu,
        Dianhui Chu,
        and~Dianbo Sui.
\IEEEcompsocitemizethanks{
\IEEEcompsocthanksitem Deyuan Liu, Zhanyue Qin, Zecheng Wang, Zhiying Tu, Dianhui Chu, and Dianbo Sui are with Harbin Institute of Technology, Harbin, China.\\
E-mail: 2022211994@stu.hit.edu.cn; 2021211875@stu.hit.edu.cn; 22s130467@stu.hit.edu.cn; tzy\_hit@hit.edu.cn; chudh@hit.edu.cn; suidianbo@hit.edu.cn\\ 
\IEEEcompsocthanksitem Hairu Wang is with University of Science and Technology of China, Hefei, China.\\
E-mail: hrwang00@mail.ustc.edu.cn\\
\IEEEcompsocthanksitem Zhao Yang is with Institute of Automation, Chinese Academy of Sciences, Beijing, China.\\
E-mail: zhao.yang@nlpr.ia.ac.cn\\
\IEEEcompsocthanksitem Fangying Rong is with Shandong Agricultural University, Shandong, China.\\
E-mail: rongfangying@gmail.com\\
\IEEEcompsocthanksitem Qingbin Liu, Xi Chen are with Tencent Inc., Shenzhen, China.\\
E-mail: qingbinliu@tencent.com; marshao@tencent.com; ryanbli@tencent.com; jasonxchen@tencent.com.\\
\IEEEcompsocthanksitem Cunhang Fan, Zhao Lv are with Anhui University, Anhui, China.\\
E-mail: cunhang.fan@ahu.edu.cn; kjlz@ahu.edu.cn.\\
\IEEEcompsocthanksitem $^{\ast}$These authors contributed equally to this work.\\
\IEEEcompsocthanksitem{Dianbo Sui is the corresponding author.} \\
}
\thanks{This work has been submitted to the IEEE for possible publication. Copyright may be transferred without notice, after which this version may no longer be accessible.}}

\markboth{Journal of \LaTeX\ Class Files,~Vol.~14, No.~8, October~2024}%
{Liu \MakeLowercase{\textit{et al.}}: Pruning via Merging: Compressing LLMs via Manifold Alignment Based Layer Merging}

\IEEEpubidadjcol

\IEEEtitleabstractindextext
{
\begin{abstract}
While large language models (LLMs) excel in many domains, their complexity and scale challenge deployment in resource-limited environments. Current compression techniques, such as parameter pruning, often fail to effectively utilize the knowledge from pruned parameters. To address these challenges, we propose Manifold-Based Knowledge Alignment and Layer Merging Compression (MKA), a novel approach that uses manifold learning and the Information Bottleneck (IB) measure to merge similar layers, reducing model size while preserving essential performance. We evaluate MKA on multiple benchmark datasets and various LLMs. Our findings show that MKA not only preserves model performance but also achieves substantial compression ratios, outperforming traditional pruning methods. Moreover, when coupled with quantization, MKA delivers even greater compression. Specifically, on the MMLU dataset using the Llama3-8B model, MKA achieves a compression ratio of 43.75\% with a minimal performance decrease of only 2.82\%. The proposed MKA method offers a resource-efficient and performance-preserving model compression technique for LLMs. We make our code available at \url{https://github.com/SempraETY/Pruning-via-Merging}
\end{abstract}

\begin{IEEEkeywords}
Model Compression, Layer Merging, Manifold Learning, Large Language Models (LLMs)
\end{IEEEkeywords}
}

\maketitle
\IEEEpeerreviewmaketitle

\IEEEdisplaynontitleabstractindextext

\IEEEraisesectionheading{\section{Introduction}\label{sec:introduction}}

\IEEEPARstart{L}{arge} Language Models (LLMs), such as GPT-4~\cite{openai2024gpt4}, Llama-3~\cite{dubey2024llama3herdmodels}, Llama-2 \cite{touvron2023llama} and Mistral \cite{jiang2024mixtral}, have demonstrated remarkable proficiency in language understanding and generation. These models, with billions of parameters trained on trillions of tokens, can handle complex tasks and exhibit emergent abilities \cite{brown2020language,chowdhery2023palm}. While these models have achieved unprecedented success, their growing complexity and scale have brought to the fore significant challenges in terms of computational resources, memory requirements, and energy consumption \cite{bender2021dangers, bommasani2021opportunities}, raising concerns about their sustainability.

To mitigate these challenges, 
researchers have developed various model compression techniques in LLM to reduce its parameter size while preserving performance \cite{Cheng2017survey, deng2020model,ganesh2021compressing,zhu2023survey,10.1145/3639364}. These techniques can be roughly categorized into two main mainstreams~\cite{men2024shortgpt}:
quantization~\cite{gholami2021survey,li2024evaluating,dettmers2022llmint8,gong2024makes,li2024evaluating} and pruning~\cite{NIPS1989_6c9882bb, han2016deep,gupta2022compression,ma2023llmpruner}. Quantization based methods aid in the reduction of the memory consumption of weights, activations, and KV caches by using the low-precision values with fewer bits
instead of the high-precision values. However, the acceleration benefits of quantization are seriously dependent on hardware support~\cite{tao-etal-2023-structured} and sometimes require additional fine-tuning to maintain performance~\cite{dettmers2023qlora,men2024shortgpt}. Compared to quantization, pruning,  especially structural pruning~\cite{li2017pruning}, eliminates redundant LLM's parameters to decrease the overall parameter count, and can be applied directly to a trained LLM without retraining and is generally more hardware-friendly than quantization approaches. While effective, pruning usually risks losing valuable model structures and determining how to prune the LLM with minimal disruption to the origin remains an unsolved problem~\cite{ma2023llm}.

\begin{figure*}[!t] 
    \centering 
    \includegraphics[width=1\textwidth]{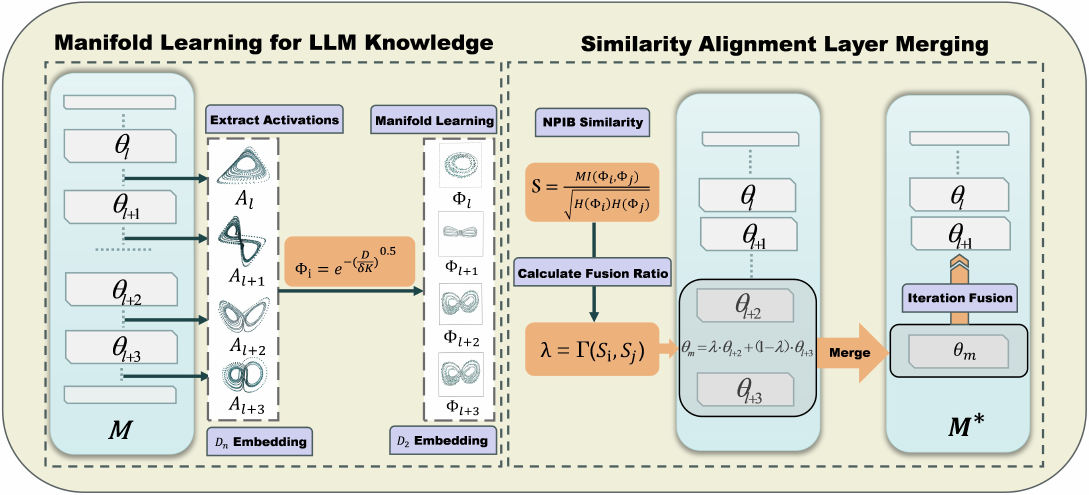}
    \caption{Manifold-Based Knowledge Alignment and Layer Merging (MKA) framework consists of two main components: (1) The left side illustrates manifold learning for LLM knowledge extraction, where layer activations are transformed into low-dimensional manifolds using the Diffusion Kernel algorithm. (2) The right side depicts the similarity-based layer merging process, employing the IB metric to identify layers with aligned knowledge.}
    \label{fig:mka_overview}
\end{figure*}
To tackle this issue head-on, we delve into the realm of model merging~\cite{wortsman2022model}, a powerful technique that seamlessly weaves together the strengths and knowledge of multiple models, creating a robust and efficient aggregation. This technique, through averaging the weights of multiple models with the same architecture, can retain essential features without significant additional resources~\cite{liu2024checkpoint, wan2024knowledge}. Furthermore, by offsetting the biases and errors of individual models, model merging often leads to greatly improved performance \cite{li2023deep}. Additional, the number of models in the merging process can be gradually and naturally reduced. However, such a useful technology are limited to merging between models currently, and few studies pay attention on merging the same internal structures within a model. 

This raises the question of whether model compression could be achieved by reducing the total number of layers through the progressive aggregation of knowledge between layers. To answer this question, we introduce Manifold-Based Knowledge Alignment and Layer Merging Compression (MKA) in this paper. MKA combines manifold learning and layer merging to preserve essential information while significantly reducing LLM parameter size. As illustrated in Figure~\ref{fig:mka_overview}, our method mainly comprises two primary components:

\textit{\textbf{Manifold Learning for LLM Knowledge:}} We employ manifold learning techniques to align knowledge across layers by extracting layer activations from a LLM and applying the Diffusion Kernel algorithm \cite{tenenbaum2000global} to learn low-dimensional manifold representations. This approach captures the nonlinear structure in the activation and achieves dimensionality reduction while preserving important activation features, enabling more effective comparison of knowledge patterns across different layers.

\textit{\textbf{Similarity Alignment Layer Merging:}} Following manifold learning, we use the Information Bottleneck (IB) measure \cite{tishby2000information} to construct a similarity matrix that quantifies the similarity between layers by maximizing their mutual information while considering the entropy of each layer. Based on this similarity matrix, we select the most similar layer pairs for merging.

To rigorously validate the effectiveness of MKA, we conduct extensive empirical evaluations on a diverse array of benchmark datasets, like MMLU and PIQA,  and a wide range of state-of-the-art large language models, including Llama-3 series with 8B and 70B parameters, Llama-2 series with 7B and 13B parameters, and Mixtral-7B. Our experimental results indicate that MKA can maintain good performance while achieving a significant compression ratio, outperforming existing pruning methods and achieving even greater compression when combined with quantization. For example, on the MMLU dataset with Llama3-8B, MKA can achieve a compression ratio of 43.75\% with only a 2.82\% performance drop.

In summary, the main contributions of this paper are as follows:

\begin{itemize}[noitemsep, topsep=0pt, left=3pt]
\item We introduce MKA, an innovative model compression technique that leverages manifold learning to align and integrate knowledge across layers, achieving significant reductions in model size while preserving performance.
\item We develop a manifold-based knowledge alignment approach, utilizing the Diffusion Kernel and Information Bottleneck (IB) to effectively capture and align similarities between layers in the parameter space.
\item We validate the efficacy of MKA through comprehensive experiments on multiple benchmark datasets and a variety of large language models, demonstrating its capability to achieve substantial compression without compromising model performance.
\end{itemize}

\section{Related Work}

Our proposed Manifold-Based Knowledge Alignment and Layer Merging (MKA) framework builds upon and integrates several key areas of research in deep model optimization for efficiency. In the following subsections, we first explore methods for learning high-dimensional data, which are essential for understanding the complex structures within large language models. We then review the primary model compression techniques, including quantization, pruning, and knowledge distillation, which aim to reduce the computational and memory footprint of these models. Finally, we delve into the emerging field of model merging, highlighting recent advancements that inform our approach to aggregating and aligning knowledge across model layers. 

\subsection{Methods for Learning High-Dimensional Data.}
High-dimensional data poses significant challenges in machine learning, primarily due to the curse of dimensionality, which can lead to increased computational complexity and overfitting \cite{hastie2009elements}. To address these challenges, manifold learning techniques have been extensively developed to capture the underlying low-dimensional structures within high-dimensional datasets. Techniques such as Principal Component Analysis (PCA) \cite{jolliffe2002principal}, t-Distributed Stochastic Neighbor Embedding (t-SNE) \cite{van2008visualizing}, and Uniform Manifold Approximation and Projection (UMAP) \cite{mcinnes2018umap} have been widely adopted for dimensionality reduction and data visualization. More recently, diffusion-based methods like Diffusion Maps \cite{coifman2005geometric} and the Diffusion Kernel algorithm \cite{tenenbaum2000global} have shown promise in preserving the intrinsic geometry of data while facilitating efficient computation. These manifold learning approaches are crucial for understanding and aligning complex patterns in high-dimensional spaces, making them well-suited for applications in large language model (LLM) compression and knowledge alignment.

\subsection{Model Compression Methods.}
Model compression has emerged as a vital area of research aimed at reducing the computational and memory footprint of large-scale models without significantly compromising their performance \cite{Cheng2017survey, deng2020model, ganesh2021compressing, zhu2023survey, 10.1145/3639364}. The primary techniques in model compression include quantization, pruning, and knowledge distillation. 

\textbf{Quantization} involves reducing the precision of the model's weights and activations from high-bit representations (e.g., 32-bit floating-point) to lower-bit formats (e.g., 8-bit integers), thereby decreasing memory usage and accelerating inference \cite{gholami2021survey, li2024evaluating, dettmers2022llmint8, gong2024makes}. While effective, quantization often requires specialized hardware support and may necessitate additional fine-tuning to maintain model accuracy \cite{dettmers2023qlora, men2024shortgpt}.

\textbf{Pruning} techniques focus on eliminating redundant or less significant parameters from the model, thereby reducing the overall parameter count and computational requirements \cite{NIPS1989_6c9882bb, han2016deep, gupta2022compression, ma2023llmpruner}. Pruning can be categorized into unstructured pruning, which removes individual weights, and structured pruning, which removes entire neurons or channels \cite{li2017pruning}. Structural pruning is particularly advantageous for its hardware friendliness and the ability to directly apply it to pre-trained models without extensive retraining \cite{ma2023llm}.

\textbf{Knowledge Distillation} involves training a smaller "student" model to replicate the behavior of a larger "teacher" model, thereby transferring knowledge while reducing model size \cite{hinton2015distilling}. This approach has been effective in maintaining performance levels while achieving significant compression.

Recent advancements have also explored hybrid approaches that combine multiple compression techniques to leverage their complementary strengths \cite{men2024shortgpt}. These methods aim to achieve higher compression ratios and better performance retention by integrating quantization, pruning, and distillation strategies.

\subsection{Model Merging.}
Model merging is a burgeoning area that focuses on combining multiple models to create a single, more robust model by leveraging the strengths and knowledge of its constituents \cite{wortsman2022model}. Traditional approaches, such as Model Soup \cite{wortsman2022model}, utilize simple weight averaging to merge models with the same architecture, effectively blending their learned representations. However, this method can be limited by the need for identical architectures and the potential for performance degradation if the individual models are not sufficiently aligned.

Advancements in model merging have introduced more sophisticated techniques to enhance the robustness and performance of the merged models. Checkpoint Merging \cite{liu2024checkpoint} employs Bayesian optimization to selectively weight and integrate different model checkpoints, resulting in a more stable and high-performing merged model. Similarly, MindMerger \cite{huang2024mindmerger} facilitates the fusion of models with varying specializations, thereby enhancing the overall capabilities of the resultant model by integrating diverse knowledge bases.

Dynamic expert merging methods, such as DELLA-Merging \cite{2024arXiv240611617T}, dynamically incorporate specialized expert models during inference, allowing the merged model to adapt to a wide range of tasks. Adaptive weighting approaches like AdaMerging \cite{yang2023adamerging} and MetaGPT \cite{2024arXiv240611385Z} leverage meta-learning and dynamic weighting schemes to fine-tune the merging process, ensuring optimal integration of the strengths of constituent models.

Furthermore, task-oriented merging strategies, including Task Arithmetic \cite{ilharco2022editing}, Language and Task Arithmetic \cite{chronopoulou2023language}, and Task Arithmetic in Tangent Space \cite{ortiz2024task}, focus on blending models trained on different tasks to create versatile LLMs capable of handling multiple applications. These approaches enhance the flexibility and applicability of merged models, making them more adaptable to diverse real-world scenarios.

Despite the progress in model merging, most existing methods focus on merging distinct models rather than addressing the internal layer structures within a single model. Our proposed Manifold-Based Knowledge Alignment and Layer Merging (MKA) framework addresses this gap by enabling the compression of LLMs through the progressive aggregation of knowledge across layers, thereby reducing the total number of layers while preserving essential model performance.

\section{Preliminaries}
\label{sec:preliminaries}

In this section, we introduce the foundational concepts and theoretical frameworks essential for understanding the proposed Manifold-Based Knowledge Alignment and Layer Merging Compression (MKA) method. We begin by discussing overparameterization and redundancy in Large Language Models (LLMs), then present the manifold hypothesis in the context of neural representations and diffusion geometry, and finally introduce mutual information within the Information Bottleneck (IB) framework.

\subsection{Overparameterization and Redundancy in Large Language Models}

Let \(\mathcal{D} = \{(\mathbf{x}_i, y_i)\}_{i=1}^N\) denote a dataset, where \(\mathbf{x}_i \in \mathcal{X}\) are input sequences and \(y_i \in \mathcal{Y}\) are the corresponding targets. A Large Language Model (LLM) is modeled as a parameterized function \(f_{\boldsymbol{\theta}}: \mathcal{X} \rightarrow \mathcal{Y}\) with parameters \(\boldsymbol{\theta} \in \mathbb{R}^d\), where \(d\) is the total number of parameters. An LLM is said to be \emph{overparameterized} if \(d \gg N\).
Overparameterization often results in redundancy; that is, different parameter configurations yield similar functional mappings. Formally, for a given \(\epsilon > 0\) and small constant \(\delta\), a subset of parameters \(\mathcal{S} \subset \boldsymbol{\theta}\) is considered \emph{redundant} if there exists an alternative parameter set \(\boldsymbol{\theta}'\) satisfying
\begin{align}
\|\boldsymbol{\theta}' - \boldsymbol{\theta}\| \leq \epsilon \quad \text{and} \quad L(f_{\boldsymbol{\theta}'}, \mathcal{D}) \leq L(f_{\boldsymbol{\theta}}, \mathcal{D}) + \delta.
\end{align}
This observation motivates the exploration of layer merging as a strategy for model compression.

\subsection{Manifold Hypothesis in Neural Representations and Diffusion Geometry}

The \emph{manifold hypothesis} posits that high-dimensional data encountered in real-world applications lie on a low-dimensional manifold embedded in the ambient space \(\mathbb{R}^D\). A manifold \(\mathcal{M}\) of dimension \(k\) is a topological space such that for every point \(p \in \mathcal{M}\), there exists a neighborhood \(U\) and a homeomorphism \(\phi: U \rightarrow \mathbb{R}^k\).

In the context of neural networks, let \(\mathbf{H}^l_i \in \mathbb{R}^{d_l}\) denote the activation of layer \(l\) for input \(\mathbf{x}_i\). The collection \(\{\mathbf{H}^l_i\}_{i=1}^N\) is assumed to lie on a manifold \(\mathcal{M}_l\) of intrinsic dimension \(k_l\) (with \(k_l \ll d_l\)) embedded in \(\mathbb{R}^{d_l}\). This assumption is supported by the observation that deep networks implicitly perform dimensionality reduction.

To analyze the intrinsic geometry of the activation manifold, we utilize diffusion geometry. Consider a graph \(\mathcal{G} = (\mathcal{V}, \mathcal{E})\) where each node corresponds to an activation vector and the edge weights are defined via a Gaussian kernel. Specifically, for two activation vectors \(\mathbf{H}^l_i\) and \(\mathbf{H}^l_j\), the affinity is given by
\begin{align}
W_{ij} = \exp\left(-\frac{\|\mathbf{H}^l_i - \mathbf{H}^l_j\|^2}{\sigma^2}\right),
\end{align}
where \(\sigma > 0\) controls the neighborhood scale. The degree matrix \(D\) is defined by
\begin{align}
D_{ii} = \sum_{j=1}^N W_{ij},
\end{align}
and the diffusion operator is then
\begin{align}
P = D^{-1} W.
\end{align}
The operator \(P\) captures the connectivity of the data manifold by serving as the transition probability matrix for a random walk on \(\mathcal{G}\).

The diffusion map is constructed via the spectral decomposition of \(P\). Suppose that \(P\) is reversible (which follows from the symmetry of \(W\)) and let \(\{(\lambda_j, \phi_j)\}_{j=1}^N\) be its eigenpairs. The diffusion map at time \(t\) is defined as
\begin{align}
\Phi_t(i) = \left( \lambda_1^t \phi_1(i), \lambda_2^t \phi_2(i), \dots, \lambda_k^t \phi_k(i) \right).
\end{align}
In many cases, the first eigenvector corresponding to \(\lambda_1=1\) is omitted to focus on the nontrivial geometry; the indexing may be adjusted accordingly.

\subsection{Mutual Information and the Information Bottleneck Principle}

Mutual information is used to quantify the similarity between representations. For a continuous random variable \(\mathbf{X}\) with density \(p(\mathbf{x})\), the differential entropy is defined as
\begin{align}
H(\mathbf{X}) = -\int p(\mathbf{x}) \log p(\mathbf{x}) \, d\mathbf{x}.
\end{align}
The mutual information between \(\mathbf{X}\) and \(\mathbf{Y}\) is given by
\begin{align}
I(\mathbf{X}; \mathbf{Y}) = H(\mathbf{X}) + H(\mathbf{Y}) - H(\mathbf{X}, \mathbf{Y}).
\end{align}
Within our framework, mutual information serves as a metric for comparing the diffusion map embeddings of different layers.

The Information Bottleneck (IB) principle provides a framework for obtaining a compressed representation that retains relevant information. For random variables \(\mathbf{X}\) and \(\mathbf{Y}\), the IB objective is to find a mapping \(p(\mathbf{T}|\mathbf{X})\) that minimizes
\begin{align}
\min_{p(\mathbf{T}|\mathbf{X})} \; I(\mathbf{X}; \mathbf{T}) - \beta I(\mathbf{T}; \mathbf{Y}),
\end{align}
where \(\beta > 0\) balances compression against the preservation of relevant information. In our context, the goal is to merge the diffusion map embeddings from two layers into a single representation that maximizes mutual information with the target variable \(\mathbf{Y}\) while minimizing redundancy.

\section{Manifold Learning for Internal Representations}
\label{sec:manifold_learning}

In this section, we introduce the Manifold-Based Knowledge Alignment (MKA) framework, which leverages the manifold hypothesis and diffusion geometry to identify and merge redundant layers within large language models (LLMs). Building on the concepts outlined in Section~\ref{sec:preliminaries}, we describe the extraction of high-dimensional activations, the construction of the diffusion operator, the spectral decomposition leading to diffusion maps, and the subsequent alignment and merging of layers using information-theoretic measures.

\subsection{Extraction of High-Dimensional Activations}
\label{subsec:activation_extraction}

Let $\mathcal{M}$ denote an LLM processing a dataset $\mathcal{D} = \{\mathbf{x}_i\}_{i=1}^N$. For each input $\mathbf{x}_i$, the model produces activations at each layer. Specifically, we denote by $\mathbf{H}^l_i \in \mathbb{R}^{d_l}$ the activation of layer $l$ for input $\mathbf{x}_i$, where $d_l$ is the activation dimension at layer $l$. The activations are computed via
\begin{align}
\mathbf{H}^0_i &= \mathrm{Embed}(\mathbf{x}_i), \\
\mathbf{H}^l_i &= f_{\boldsymbol{\theta}_l}(\mathbf{H}^{l-1}_i), \quad l = 1,2,\dots,L,
\end{align}
where $f_{\boldsymbol{\theta}_l}$ denotes the transformation associated with layer $l$, parameterized by $\boldsymbol{\theta}_l$. The collection $\{\mathbf{H}^l_i\}_{i=1}^N$ thus forms a high-dimensional dataset intrinsic to each layer.

\subsection{Construction of the Diffusion Operator}
\label{subsec:diffusion_operator}

To analyze the manifold structure underlying these activations, we construct a weighted graph $\mathcal{G}_l = (\mathcal{V}_l,\mathcal{E}_l)$ for each layer $l$, where each node corresponds to an activation vector $\mathbf{H}^l_i$. The edges are weighted according to the affinity between activation vectors.

Given two activation vectors $\mathbf{H}^l_i$ and $\mathbf{H}^l_j$, we define the affinity using a Gaussian kernel:
\begin{align}
W_{ij} = K(\mathbf{H}^l_i,\mathbf{H}^l_j) = \exp\left(-\frac{\|\mathbf{H}^l_i - \mathbf{H}^l_j\|^2}{\sigma^2}\right),
\end{align}
where $\sigma>0$ is a bandwidth parameter that controls the local neighborhood scale.

The degree matrix $D$ is diagonal with entries
\begin{align}
D_{ii} = \sum_{j=1}^{N}W_{ij}.
\end{align}

We then define the diffusion operator as the normalized affinity matrix:
\begin{align}
P = D^{-1}W.
\end{align}
This operator governs the transition probabilities of a random walk on the graph $\mathcal{G}_l$ and thus captures the intrinsic geometry of the activation manifold.

\subsection{Spectral Decomposition and Diffusion Maps}
\label{subsec:spectral_decomposition}

The operator $P$ encodes the manifold geometry, and its spectral properties allow us to construct diffusion maps—lower-dimensional embeddings that preserve the manifold’s structure.

Assume that the affinity matrix $W$ is symmetric and positive semidefinite and that $P$ is reversible (which is the case when the underlying Markov chain is reversible). Although $P$ is not symmetric, it is similar to the symmetric matrix
\begin{align}
\widetilde{P} = D^{-1/2} W D^{-1/2}.
\end{align}
Since $\widetilde{P}$ is symmetric, it has a complete set of real eigenvalues $\{\lambda_k\}_{k=1}^N$ and corresponding orthonormal eigenvectors $\{\boldsymbol{\phi}_k\}_{k=1}^N$, so that
\begin{align}
P\boldsymbol{\phi}_k = \lambda_k \boldsymbol{\phi}_k.
\end{align}
Without loss of generality, we order the eigenvalues as
\begin{align}
1 = \lambda_1 \geq \lambda_2 \geq \dots \geq \lambda_N \geq -1.
\end{align}
The largest eigenvalue is $\lambda_1=1$, and its corresponding eigenvector is constant.

Using the spectral decomposition, the diffusion map at time $t$ is defined as the mapping
\begin{align}
\Phi_t(\mathbf{H}^l_i) = \Bigl(\lambda_2^t\,\phi_2(i),\, \lambda_3^t\,\phi_3(i),\, \dots,\, \lambda_{k+1}^t\,\phi_{k+1}(i)\Bigr),
\end{align}
where $k$ is the target embedding dimension and $\phi_j(i)$ denotes the $i$th component of $\boldsymbol{\phi}_j$. (Note that the trivial eigenvector associated with $\lambda_1$ is omitted.)

\subsection{Preservation of Manifold Structure}
\label{subsec:manifold_preservation}

The diffusion map is designed to preserve the intrinsic geometry of the activation manifold by capturing multi-scale connectivity. In particular, the Euclidean distance in the diffusion space approximates the diffusion distance between points.

For any two activation vectors $\mathbf{H}^l_i$ and $\mathbf{H}^l_j$, the diffusion distance is defined by
\begin{align}
D_t(i,j)^2 = \sum_{k=2}^{k+1}\lambda_k^{2t}\,\bigl(\phi_k(i)-\phi_k(j)\bigr)^2.
\end{align}
By the very definition of the diffusion map,
\begin{align}
\|\Phi_t(\mathbf{H}^l_i)-\Phi_t(\mathbf{H}^l_j)\|_2^2 = \sum_{k=2}^{k+1}\lambda_k^{2t}\,\bigl(\phi_k(i)-\phi_k(j)\bigr)^2,
\end{align}
so that the Euclidean distance in the embedding space approximates the diffusion distance. Moreover, as $t\to\infty$, the influence of the smaller eigenvalues diminishes, and the diffusion map increasingly emphasizes the global structure of the manifold.

\subsection{Layer Similarity Measure via Mutual Information}
\label{subsec:mutual_information_similarity}

To quantify the similarity between representations in different layers, we compare their diffusion map embeddings. Let
\begin{align}
\mathbf{\Psi}^l = \Phi_t(\mathbf{H}^l) \quad \text{and} \quad \mathbf{\Psi}^m = \Phi_t(\mathbf{H}^m)
\end{align}
be the diffusion map embeddings for layers $l$ and $m$, respectively. We define their similarity in terms of the mutual information (MI)
\begin{align}
I(\mathbf{\Psi}^l;\mathbf{\Psi}^m) = H(\mathbf{\Psi}^l) + H(\mathbf{\Psi}^m) - H(\mathbf{\Psi}^l,\mathbf{\Psi}^m),
\end{align}
where $H(\cdot)$ denotes differential entropy. In order to simplify the calculation, we assume that $\mathbf{\Psi}^l$ and $\mathbf{\Psi}^m$ are jointly Gaussian random variables. Under this joint Gaussianity assumption, the MI can be written in closed form as
\begin{align}
I(\mathbf{\Psi}^l;\mathbf{\Psi}^m) = \frac{1}{2}\ln\!\left(\frac{|\Sigma_{\mathbf{\Psi}^l}||\Sigma_{\mathbf{\Psi}^m}|}{|\Sigma_{\mathbf{\Psi}^l,\mathbf{\Psi}^m}|}\right),
\end{align}
where $\Sigma_{\mathbf{\Psi}^l}$ and $\Sigma_{\mathbf{\Psi}^m}$ denote the covariance matrices of $\mathbf{\Psi}^l$ and $\mathbf{\Psi}^m$, respectively, and $\Sigma_{\mathbf{\Psi}^l,\mathbf{\Psi}^m}$ is their joint covariance matrix.

To facilitate comparison across different layer pairs, we further define the \emph{normalized mutual information} (NMI) as
\begin{align}
S_{lm} = \frac{I(\mathbf{\Psi}^l;\mathbf{\Psi}^m)}{\sqrt{H(\mathbf{\Psi}^l)H(\mathbf{\Psi}^m)}}.
\end{align}
This normalization constrains $S_{lm}$ to lie in a consistent range, thereby enabling meaningful similarity assessments between layers.

\subsection{Manifold-Based Knowledge Alignment and Layer Merging Compression (MKA)}
\label{subsec:layer_merging_algorithm}

Algorithm~\ref{algorithm} summarizes the MKA procedure. In brief, for each layer the activations are first extracted and embedded using the diffusion map. Next, pairwise similarities between layers are computed via their mutual information. Finally, when the similarity score between a pair of layers exceeds a predetermined threshold $\tau$, the layers are merged according to a weighted combination of their parameters.

\begin{algorithm*}[t]
\caption{Manifold-Based Knowledge Alignment and Layer Merging Compression (MKA)}
\label{algorithm}
\begin{algorithmic}[1]
\REQUIRE LLM $\mathcal{M}$ with layers $\{L_1,L_2,\dots,L_N\}$ and parameters $\Theta=\{\theta_1,\theta_2,\dots,\theta_N\}$, dataset $\omega$
\ENSURE Compressed model $\mathcal{M}^*$ with aligned representations
\STATE Extract activations $\mathcal{H} = \{\mathbf{H}^l_i\}$ for each layer $l$ on dataset $\omega$
\FOR{each layer $l$}
    \STATE Compute pairwise distances among activations $\{\mathbf{H}^l_i\}$
    \STATE Construct the affinity matrix $W^{(l)}$ using a Gaussian kernel with bandwidth $\sigma$
    \STATE Compute the diffusion map embedding $\mathbf{\Psi}^l$ via the eigendecomposition of $W^{(l)}$
\ENDFOR
\FOR{each pair of layers $(l,m)$}
    \STATE Estimate the covariance matrices $\Sigma_{\mathbf{\Psi}^l}$, $\Sigma_{\mathbf{\Psi}^m}$, and joint covariance $\Sigma_{\mathbf{\Psi}^l,\mathbf{\Psi}^m}$
    \STATE Compute $I(\mathbf{\Psi}^l;\mathbf{\Psi}^m)$ and the normalized similarity score $S_{lm}$
\ENDFOR
\WHILE{there exists a pair $(l,m)$ with $S_{lm}\ge\tau$}
    \STATE Determine the merging weight $\alpha$, for example via $\alpha = S_{lm}$ (or, alternatively, using a softmax formulation)
    \STATE Merge the parameters: $\widetilde{\theta}_c = \alpha\,\theta_l + (1-\alpha)\,\theta_m$
    \STATE Replace layers $l$ and $m$ with the merged layer using $\widetilde{\theta}_c$ and update $\mathcal{M}$ accordingly
\ENDWHILE
\RETURN Compressed model $\mathcal{M}^*$
\end{algorithmic}
\end{algorithm*}

\subsection{Layer Merging via the Information Bottleneck Principle}
\label{subsec:layer_merging_ib}

To guide the layer merging process, we adopt an information-theoretic perspective based on the Information Bottleneck (IB) principle. The IB framework seeks to extract a compressed representation that preserves the information most relevant to a target variable.

In its standard form, given random variables $\mathbf{X}$ and $\mathbf{Y}$, the IB objective is to find a mapping $p(\mathbf{T}|\mathbf{X})$ that minimizes
\begin{align}
\mathcal{L}_{\mathrm{IB}} = I(\mathbf{X};\mathbf{T}) - \beta\,I(\mathbf{T};\mathbf{Y}),
\end{align}
where $\beta>0$ balances the trade-off between compression (minimizing $I(\mathbf{X};\mathbf{T})$) and prediction (maximizing $I(\mathbf{T};\mathbf{Y})$).

For merging layers $l$ and $m$, we set $\mathbf{X}=(\mathbf{\Psi}^l,\mathbf{\Psi}^m)$ and seek a compressed representation $\mathbf{\Psi}^c$ that captures their shared information. The corresponding IB objective is
\begin{align}
\mathcal{L}_{\mathrm{IB}} = I\bigl((\mathbf{\Psi}^l,\mathbf{\Psi}^m);\mathbf{\Psi}^c\bigr) - \beta\,I(\mathbf{\Psi}^c;\mathbf{Y}).
\end{align}
To make the optimization tractable, we restrict the mapping to a deterministic linear combination:
\begin{align}
\mathbf{\Psi}^c = \alpha\,\mathbf{\Psi}^l + (1-\alpha)\,\mathbf{\Psi}^m, \quad \alpha\in[0,1].
\end{align}

Under the joint Gaussian assumption, the covariance of $\mathbf{\Psi}^c$ is given by
\begin{align}
\Sigma_{\mathbf{\Psi}^c} = \alpha^2\,\Sigma_{\mathbf{\Psi}^l} + (1-\alpha)^2\,\Sigma_{\mathbf{\Psi}^m} + 2\alpha(1-\alpha)\,\Sigma_{\mathbf{\Psi}^l,\mathbf{\Psi}^m},
\end{align}
and if $\Sigma_{\mathbf{\Psi}^c,\mathbf{Y}}$ denotes the cross-covariance between $\mathbf{\Psi}^c$ and $\mathbf{Y}$, then
\begin{align}
\Sigma_{\mathbf{\Psi}^c,\mathbf{Y}} = \alpha\,\Sigma_{\mathbf{\Psi}^l,\mathbf{Y}} + (1-\alpha)\,\Sigma_{\mathbf{\Psi}^m,\mathbf{Y}}.
\end{align}
The conditional covariance is defined as
\begin{align}
\Sigma_{\mathbf{\Psi}^c|\mathbf{Y}} = \Sigma_{\mathbf{\Psi}^c} - \Sigma_{\mathbf{\Psi}^c,\mathbf{Y}}\,\Sigma_{\mathbf{Y}}^{-1}\,\Sigma_{\mathbf{Y},\mathbf{\Psi}^c}.
\end{align}

Since $\mathbf{\Psi}^c$ is a deterministic function of $(\mathbf{\Psi}^l,\mathbf{\Psi}^m)$ (i.e., $H(\mathbf{\Psi}^c|(\mathbf{\Psi}^l,\mathbf{\Psi}^m))=0$), the mutual information terms reduce to differential entropies. In particular, using the standard Gaussian formula for differential entropy,
\begin{align}
H(\mathbf{\Psi}^c) &= \frac{1}{2}\ln\Bigl((2\pi e)^{d_c}|\Sigma_{\mathbf{\Psi}^c}|\Bigr), \\
H(\mathbf{\Psi}^c|\mathbf{Y}) &= \frac{1}{2}\ln\Bigl((2\pi e)^{d_c}|\Sigma_{\mathbf{\Psi}^c|\mathbf{Y}}|\Bigr),
\end{align}
the IB objective simplifies to
\begin{align}
\mathcal{L}_{\mathrm{IB}} = \frac{1}{2}\Bigl[(1-\beta)\ln|\Sigma_{\mathbf{\Psi}^c}| + \beta\,\ln|\Sigma_{\mathbf{\Psi}^c|\mathbf{Y}}|\Bigr] + \mathrm{const},
\end{align}
where the constant term is independent of $\alpha$.

In principle, one may optimize $\mathcal{L}_{\mathrm{IB}}$ with respect to $\alpha$ by differentiating
\begin{align}
\frac{\partial \mathcal{L}_{\mathrm{IB}}}{\partial \alpha} = \frac{1}{2}\Biggl[(1-\beta)\,\mathrm{Tr}\Bigl(\Sigma_{\mathbf{\Psi}^c}^{-1}\frac{\partial\Sigma_{\mathbf{\Psi}^c}}{\partial \alpha}\Bigr) + \beta\,\mathrm{Tr}\Bigl(\Sigma_{\mathbf{\Psi}^c|\mathbf{Y}}^{-1}\frac{\partial\Sigma_{\mathbf{\Psi}^c|\mathbf{Y}}}{\partial \alpha}\Bigr)\Biggr]
\end{align}
and setting the derivative to zero. For example, one may compute
\begin{align}
\frac{\partial\Sigma_{\mathbf{\Psi}^c}}{\partial \alpha} = 2\alpha\,\Sigma_{\mathbf{\Psi}^l} - 2(1-\alpha)\,\Sigma_{\mathbf{\Psi}^m} + 2(1-2\alpha)\,\Sigma_{\mathbf{\Psi}^l,\mathbf{\Psi}^m}.
\end{align}
A similar expression can be derived for $\partial\Sigma_{\mathbf{\Psi}^c|\mathbf{Y}}/\partial \alpha$. In practice, however, this equation does not admit a closed-form solution. Hence, we adopt an approximate strategy by setting
\begin{align}
\alpha = S_{lm},
\end{align}
where $S_{lm}$ is the normalized mutual information defined earlier. This heuristic assigns greater weight to the layer with higher shared information, thereby approximately minimizing the IB objective while maintaining computational efficiency.

With the weight $\alpha$ determined, the parameters of layers $l$ and $m$ are merged via
\begin{align}
\widetilde{\boldsymbol{\theta}}_c = \alpha\,\boldsymbol{\theta}_l + (1-\alpha)\,\boldsymbol{\theta}_m.
\end{align}
This merged parameter set $\widetilde{\boldsymbol{\theta}}_c$ is used to replace the original layers, resulting in a compressed model $\mathcal{M}^*$.

\subsection{Impact on Model Performance}
\label{subsec:impact_model_performance}

Merging layers alters the parameter space of the model, which may in turn affect its performance. 

Let $\delta\boldsymbol{\theta} = \widetilde{\boldsymbol{\theta}}_c - \boldsymbol{\theta}$ denote the shift in parameters due to merging. Suppose that the loss function $\mathcal{L}(\boldsymbol{\theta})$ is twice differentiable and locally convex in a neighborhood around $\boldsymbol{\theta}$. A second-order Taylor expansion about $\boldsymbol{\theta}$ yields
\begin{align}
\mathcal{L}(\widetilde{\boldsymbol{\theta}}_c) \approx \mathcal{L}(\boldsymbol{\theta}) + \nabla\mathcal{L}(\boldsymbol{\theta})^\top\delta\boldsymbol{\theta} + \frac{1}{2}\,\delta\boldsymbol{\theta}^\top\nabla^2\mathcal{L}(\boldsymbol{\theta})\,\delta\boldsymbol{\theta}.
\end{align}
At a local minimum where $\nabla\mathcal{L}(\boldsymbol{\theta}) = \mathbf{0}$, the increase in loss is bounded by
\begin{align}
\Delta\mathcal{L} &= \mathcal{L}(\widetilde{\boldsymbol{\theta}}_c)-\mathcal{L}(\boldsymbol{\theta}) \nonumber\\[1mm]
&\le \frac{1}{2}\,\lambda_{\max}\,\|\delta\boldsymbol{\theta}\|^2,
\end{align}
where $\lambda_{\max}$ is the largest eigenvalue of the Hessian $\nabla^2\mathcal{L}(\boldsymbol{\theta})$. This bound follows from the Rayleigh quotient and provides a guarantee on the loss increase due to the merging operation.

\section{Experiments}

We conduct a comprehensive set of experiments to evaluate the effectiveness and generalizability of our MKA method across various domains. Moreover, we aim to compare our approach with pruning techniques to assess whether it offers improvements and to investigate if it can be combined with quantization methods to achieve even higher compression ratios.

\subsection{Experimental Setup}
\subsubsection{Datasets}
We conduct evaluations using the MKA methods across various benchmark datasets, each specifically designed to test various facets of language comprehension and generation. In detail, \textbf{MMLU}~\cite{hendrycks2020measuring}  evaluates broad language understanding across a wide range of domains. \textbf{PIQA}~\cite{bisk2020piqa}  is designed to test models on commonsense reasoning in the physical world, aiming to assess NLP models' grasp of everyday physical interactions. \textbf{HellaSwag}~\cite{zellers2019hellaswag} is a challenge dataset for commonsense natural language inference, consisting of event descriptions with multiple possible continuations, where the task is to select the most plausible one. \textbf{RACE-H}~\cite{lai2017race} is a large-scale reading comprehension dataset collected from English exams for Chinese high school students, featuring a high proportion of questions that require reasoning. \textbf{BoolQ}~\cite{clark2019boolq} is a reading comprehension dataset focusing on naturally occurring yes/no questions that often query for complex, non-factoid information and require difficult entailment-like inference to answer correctly.

\subsubsection{LLMs}
In our experiments, we employ the Llama2~\cite{touvron2023llama}, Llama3, Llama3.2, and Mistral~\cite{jiang2023mistral} models, each distinct in their capabilities and configurations: \textbf{Llama2}: Encompassing models from 7 billion to 13 billion parameters, exhibits superior performance and safety on diverse benchmarks. \textbf{Llama3}: Featuring models with 8 billion parameters, which offers state-of-the-art performance and advanced reasoning capabilities. \textbf{Llama3.2}: Featuring models with 3 billion parameters that balances performance and number of parameters. \textbf{Mistral}: We use the 7 billion parameter version of Mistral that surpasses Llama-2 and Llama-1 in performance and efficiency, leveraging grouped-query and sliding window attention mechanisms for optimal inference across lengthy sequences.

\subsubsection{Baselines}
In this study, we assess the effectiveness of our proposed method, MKA, through two distinct comparative analyses. Firstly, we evaluate MKA directly against several well-established pruning techniques to gauge its standalone efficacy in reducing model size while maintaining performance. Secondly, we extend the comparison to include scenarios where both the traditional pruning methods and MKA are further enhanced through quantization. The baseline methods included in our analysis are: 
\textbf{PruneMe}~\cite{gromov2024unreasonableineffectivenessdeeperlayers}: A pruning method identifies the optimal block of layers to prune by considering the similarity across layers. \textbf{SLEB}~\cite{song2024slebstreamliningllmsredundancy}: A pruning method designed to streamline LLMs by eliminating redundant transformer blocks. We choose the transformer block as the fundamental unit for pruning, because LLMs exhibit block-level redundancy with high similarity between the outputs of neighboring blocks. 
\textbf{Shortened}~\cite{kim2024shortenedllamadepthpruning}: A pruning method first uses a simple metric to identify unimportant blocks and then performs a simple one-shot pruning. \textbf{ShortGPT}~\cite{men2024shortgpt}: A pruning method that removes redundant layers from large language models based on a Block Influence metric, which assesses the significance of each layer. \textbf{Reverse}: A heuristic approach where the importance of layers is considered inversely proportional to their order in the model, prioritizing the retention of earlier layers. \textbf{SmoothQuant}~\cite{xiao2023smoothquant}: SmoothQuant is a training-free post-training quantization solution that enables efficient 8-bit weight and activation quantization for large language models, offering up to 1.56× speedup and 2× memory reduction with minimal accuracy loss. \textbf{GPTQ}~\cite{frantar2022gptq}: A one-shot weight quantization method that uses approximate second-order information to maintain high accuracy even with severe weight reduction. \textbf{AWQ}~\cite{lin2023awq}: A novel quantization approach that protects salient weights by adjusting per-channel scaling based on activation observations rather than weight Magnitudes.

\begin{figure*}[!t]
    \centering
    \includegraphics[width=1\textwidth]{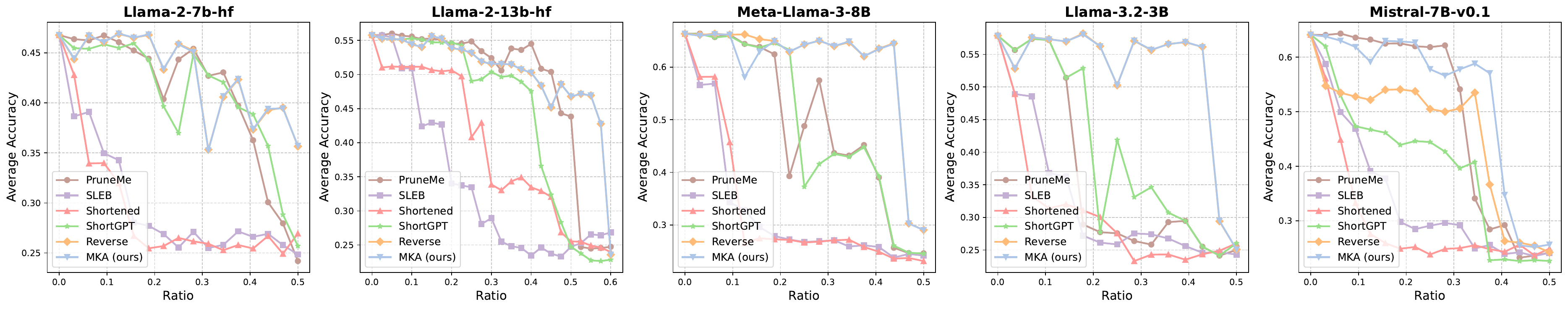}
    \caption{Performance (Accuracy) of LLMs (Llama2-7B, Llama2-13B, Llama3-8B, Llama3.2-3B, and Mistral-7B) on the MMLU dataset as the pruning ratio of various pruning methods increases.}
    \label{fig:acc}
\end{figure*}

\section{Experiments}
  
\subsection{Comparison of MKA with other structured pruning methods}

We compare the performance of MKA with baseline compression methods on the MMLU dataset using the Llama2-7B, Llama2-13B, Llama3-8B, Llama3.2-3B, and Mistral-7B models. The evaluation metric is Accuracy (ACC) during merging and pruning. The results are presented in \Cref{fig:acc}.

From \Cref{fig:acc}, we can observe that, across all models, our method improves the compression ratio while maintaining performance. Specifically, the compression ratio\footnote[3]{Note that, the compression ratio is calculated as: $\left(L_{\text{total}} - \left(\frac{L_{\text{retained}}}{Q}\right)\right) / L_{\text{total}}$, where $L_{total}$ is the total number of layers before compression, $L_{retained}$ is the number of retained layers, and $Q$ is the quantization factor.} for Llama2-7B reach 31\%, for Llama3-8B reach 44\%, for Llama3.2-3B reach 43\%, for Mistral-7B it reaches 40\%, and for Llama2-13B it reaches an impressive 58\%. Additionally, we observe several phenomena: both methods experience a collapse in model performance, but the model merging method can delay the layer collapse to some extent and stabilize the model's performance very well. Since our strategy is based on Reverse Prune, the scores for the Llama2-7B, Llama2-13B, Llama3-8B, and Llama3.2-3B models are very close to the Reverse Prune. Our hypothesis is that the pruning or merging of these models is similar, but model merging can adjust the merging ratio to surpass the effect of pruning. Moreover, for the Mistral-7B models, we noticed that the results do not closely match the Reverse Prune. 

\begin{table}[!t]
    \centering
    \normalsize
    \caption{Performance comparison of MKA and ShortGPT pruning with quantization (SmoothQuant, GPTQ, AWQ) on MMLU using Llama2-7B, Llama3-8B, and Mistral-7B. MKA outperforms ShortGPT in accuracy across all models and quantization methods at similar compression ratios with int4. The calculation of the compression ratio only considers the number of hidden layers in the model without considering the embedding layer.}
    \scalebox{0.70}{
    \begin{tabular}{c|l|c|l}
        \toprule
        \renewcommand{\arraystretch}{1.3} 
        \textbf{Model} & \textbf{Method} & \makecell{\textbf{Retained layers} \\ \textbf{(Compression Ratio)}} & \textbf{Acc} \\
        \midrule
        
        \multirow{7}{*}{\textbf{Llama2-7B}}
        & Vanilla Model& 32(0.00\%)& 46.67 \\ [2pt]
        & ShortGPT+Smooth & 16(87.50\%)& 25.67 \\ [2pt]
        & ShortGPT+GPTQ & 16(87.50\%)& 25.82 \\ [2pt]
        & ShortGPT+AWQ & 16(87.50\%)& 26.01 \\ [2pt]
        & \cellcolor{gray!30}\textbf{MKA (Ours) + Smooth} & 16(87.50\%)\cellcolor{gray!30}& \cellcolor{gray!30}\textbf{35.66 (\textcolor{red}{+9.99})}\\ [2pt]
        & \cellcolor{gray!30}\textbf{MKA (Ours) + GPTQ} & 16(87.50\%)\cellcolor{gray!30}& \cellcolor{gray!30}\textbf{35.91 (\textcolor{red}{+10.09})}\\ [2pt]
        & \cellcolor{gray!30}\textbf{MKA (Ours) + AWQ} & 16(87.50\%)\cellcolor{gray!30}& \cellcolor{gray!30}\textbf{36.23 (\textcolor{red}{+10.22})}\\ [2pt]
        \midrule
        
        \multirow{7}{*}{\textbf{Llama3-8B}}
        & Vanilla Model& 32 (0.00\%)& 66.29 \\ [2pt]
        & ShortGPT+Smooth & 18(85.94\%)& 26.54 \\ [2pt]
        & ShortGPT+GPTQ & 18(85.94\%)& 25.98 \\ [2pt]
        & ShortGPT+AWQ & 18(85.94\%)& 26.22 \\ [2pt]
        & \cellcolor{gray!30}\textbf{MKA (Ours) + Smooth} & 18(85.94\%)\cellcolor{gray!30}& \cellcolor{gray!30}\textbf{64.20 (\textcolor{red}{+37.66})}\\ [2pt]
        & \cellcolor{gray!30}\textbf{MKA (Ours) + GPTQ} & 18(85.94\%)\cellcolor{gray!30}& \cellcolor{gray!30}\textbf{62.98 (\textcolor{red}{+37.00})}\\ [2pt]
        & \cellcolor{gray!30}\textbf{MKA (Ours) + AWQ} & 18(85.94\%)\cellcolor{gray!30}& \cellcolor{gray!30}\textbf{61.66 (\textcolor{red}{+35.44})}\\ [2pt]
        \midrule
         
        \multirow{7}{*}{\textbf{Mistral-7B}}
        & Vanilla Model& 32(0.00\%)& 63.87 \\ [2pt]
        & ShortGPT+Smooth & 20(84.38\%)& 24.32 \\ [2pt]
        & ShortGPT+GPTQ & 20(84.38\%)& 23.16 \\ [2pt]
        & ShortGPT+AWQ & 20(84.38\%)& 23.96 \\ [2pt]
        & \cellcolor{gray!30}\textbf{MKA (Ours) + Smooth} & 20(84.38\%)\cellcolor{gray!30}& \cellcolor{gray!30}\textbf{56.92 (\textcolor{red}{+32.60})}\\ [2pt]
        & \cellcolor{gray!30}\textbf{MKA (Ours) + GPTQ} & 20(84.38\%)\cellcolor{gray!30}& \cellcolor{gray!30}\textbf{56.12 (\textcolor{red}{+32.96})}\\ [2pt]
        & \cellcolor{gray!30}\textbf{MKA (Ours) + AWQ} & 20(84.38\%)\cellcolor{gray!30}& \cellcolor{gray!30}\textbf{55.34 (\textcolor{red}{+31.38})}\\ [2pt] 
 
        \bottomrule 
    \end{tabular}}
    \label{tab:quant}
\end{table}

\begin{table*}[!t]
    \centering
    \caption{Comparison of different methods across MMLU, PIQA, HellaSwag, RACE-H, and BoolQ datasets at different compression ratios.}
    \normalsize
    \scalebox{0.70}{
    \begin{tabular}{c|c|c|c|c|c|c|c|c|c|c}
        \toprule
        & \multicolumn{5}{c|}{\textbf{Compression Ratio = 34.375\%}} & \multicolumn{5}{c}{\textbf{Compression Ratio = 37.5\%}} \\
        \midrule
        \textbf{Method} & \textbf{MMLU} & \textbf{PIQA} & \textbf{HellaSwag} & \textbf{RACE-H} & \textbf{BoolQ} & \textbf{MMLU} & \textbf{PIQA} & \textbf{HellaSwag} & \textbf{RACE-H} & \textbf{BoolQ} \\
        \midrule
        Vanilla Model & 66.29 & 81.12 & 74.54 & 66.07 & 66.79 & 66.29 & 81.12 & 74.54 & 66.07 & 66.79\\
        ShortGPT & 42.95 & 60.99 & 33.00 & 41.68 & 51.96 & 44.80 & 61.70 & 38.69 & 40.05 & 57.09 \\
        \cellcolor{gray!30}\textbf{MKA (Ours)} & \cellcolor{gray!30}\textbf{64.87(\textcolor{red}{+20.42})}& \cellcolor{gray!30}\textbf{67.79(\textcolor{red}{+6.80})}& \cellcolor{gray!30}\textbf{51.32(\textcolor{red}{+18.32})}& \cellcolor{gray!30}\textbf{55.20(\textcolor{red}{+13.52})}& \cellcolor{gray!30}\textbf{63.36(\textcolor{red}{+11.40})}& \cellcolor{gray!30}\textbf{62.05(\textcolor{red}{+17.25})}& \cellcolor{gray!30}\textbf{66.26(\textcolor{red}{+4.56})}& \cellcolor{gray!30}\textbf{50.16(\textcolor{red}{+11.47})}& \cellcolor{gray!30}\textbf{49.49(\textcolor{red}{+9.44})}& \cellcolor{gray!30}\textbf{63.46(\textcolor{red}{+6.37})}\\
        \bottomrule
        \addlinespace[10pt]
        \toprule
        & \multicolumn{5}{c|}{\textbf{Compression Ratio = 40.625\%}} & \multicolumn{5}{c}{\textbf{Compression Ratio = 43.75\%}} \\
        \midrule
        \textbf{Method} & \textbf{MMLU} & \textbf{PIQA} & \textbf{HellaSwag} & \textbf{RACE-H} & \textbf{BoolQ} & \textbf{MMLU} & \textbf{PIQA} & \textbf{HellaSwag} & \textbf{RACE-H} & \textbf{BoolQ} \\
        \midrule
        Vanilla Model & 66.29 & 81.12 & 74.54 & 66.07 & 66.79 & 66.29 & 81.12 & 74.54 & 66.07 & 66.79\\
        ShortGPT & 39.26 & 58.22 & 34.16 & 21.70 & 61.77 & 26.09 & 59.03 & 33.75 & 21.58 & 61.53 \\
        \cellcolor{gray!30}\textbf{MKA (Ours)} & \cellcolor{gray!30}\textbf{63.42(\textcolor{red}{+24.16})}& \cellcolor{gray!30}\textbf{65.61(\textcolor{red}{+6.25})}& \cellcolor{gray!30}\textbf{48.83(\textcolor{red}{+14.67})}& \cellcolor{gray!30}\textbf{55.26(\textcolor{red}{+33.20})}& \cellcolor{gray!30}\textbf{63.58(\textcolor{red}{+1.81})}& \cellcolor{gray!30}\textbf{64.42(\textcolor{red}{+31.31})}& \cellcolor{gray!30}\textbf{65.51(\textcolor{red}{+6.48})}& \cellcolor{gray!30}\textbf{45.10(\textcolor{red}{+11.35})}& \cellcolor{gray!30}\textbf{45.91(\textcolor{red}{+22.77})}& \cellcolor{gray!30}\textbf{62.14(\textcolor{red}{+0.51})}\\
        \bottomrule
    \end{tabular}}
    \label{tab:bench}
\end{table*}

\begin{figure*}[!t]
    \centering
    \includegraphics[width=1\textwidth]{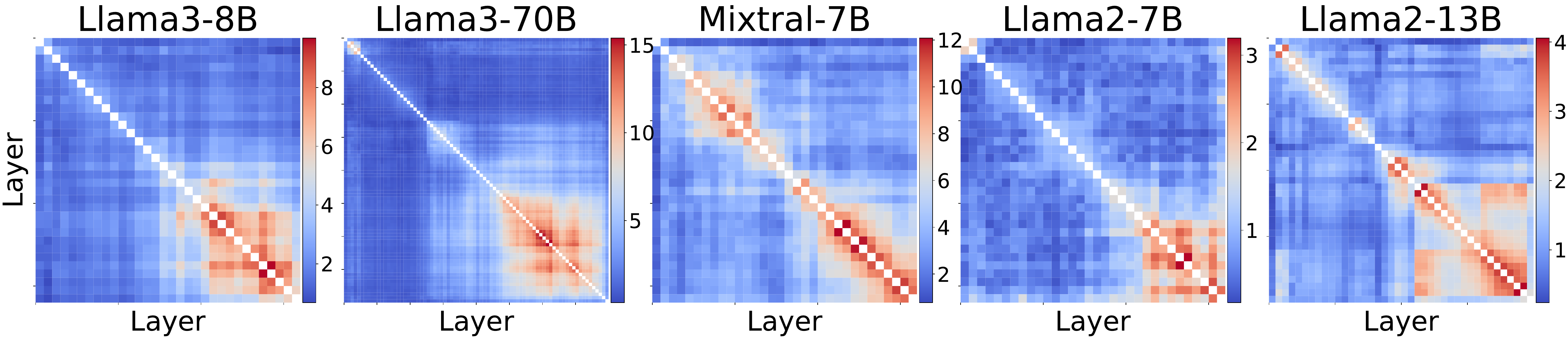}
    \caption{Similarity matrices for Llama2-7B, Llama2-13B, Llama-3-8B, Llama3.2-3B, and Mistral-7B before MKA. Later layers show high similarity, supporting layer merging.}
    \label{fig:all}
\end{figure*}

\subsection{How Does MKA Combined with Quantization Perform Compared to Pruning Combined with Quantization?}
We compare the performance of MKA with the baseline pruning method,  ShortGPT~\cite{men2024shortgpt}, on the MMLU dataset using the Llama2-7B, Llama3-8B, and Mistral-7B models. The results are shown in \Cref{tab:quant}. 

From \Cref{tab:quant}, we can see that the pruned models are able to be further quantized and maintain performance with a higher compression ratio. Notably, at a high compression ratio of around 87.50\%, MKA significantly outperforms ShortGPT. Additionally, we achieve excellent results with various quantization methods. For example, on Llama3-8B, at a compression ratio of 85.94\%, MKA with SmoothQuant achieves 64.20\%, far exceeding ShortGPT with SmoothQuant at 37.66\%. Similarly, with the GPTQ quantization method, we achieve 62.98\%, surpassing ShortGPT's 37.00\%, and with AWQ, we achieve 61.66\%, exceeding ShortGPT's 35.44\%.

\subsection{MKA vs. Other Pruning Methods on varies benchmarks}
We compared the performance of MKA and several other pruning methods on the LLama3-8B model using multiple benchmark datasets at compression ratios of 34.375\%, 37.5\%, 40.625\% and 43.75\%. The results are shown in \Cref{tab:bench}. From the results, merging can retain performance better compared to pruning. Relative to ShortGPT, our method can achieve better performance retention, with significant improvements across all datasets. For example, at a compression ratio of 34.375\% on the MMLU dataset, our method can outperform ShortGPT by 21.92\%. Similarly, on the HellaSwag dataset, our proposed method can surpass ShortGPT by 18.32\%. 

\subsection{Are Inter-Layer Knowledge Alignment Similarity Matrices Consistent Across Different Models?}
We generate layer similarity heatmaps for different models before applying MKA. These heatmaps visualize the knowledge alignment and layer merging effects of MKA on various models. Figure~\ref{fig:all} presents the similarity heatmaps for Llama2-7B, Llama2-13B, Llama-3-8B, Llama3.2-3B, and Mistral-7B models. We observe that the heatmaps for the later layers of each model exhibit high similarity values, indicating that inter-layer similarity is consistently high in the later layers across different models. This observation supports our layer merging approach. Additionally, when merging the earlier layers, we notice a collapse of the matrix in the final figure, suggesting that earlier layers have a significant influence on later layers. Thus, simple merging operations on the earlier layers of the model are not feasible.

\subsection{Iterative Nature of MKA}

It's important to note that the MKA method already incorporates an iterative process in its design. For example, when we merge layers 31 and 32, we obtain a fused layer, which is then merged with layer 30 in the next iteration. We have compared this approach with an alternative method where each layer is allowed to merge only once (e.g., merging layers 31 and 32, then 30 and 29 separately). Our experiments on the Llama3-8B model using the MMLU dataset demonstrate that MKA's iterative approach yields superior performance in terms of minimizing accuracy degradation. The results are presented in Table~\ref{tab:iterative_comparison}.

\section{Discussion}

\subsection{Extension to Multimodal and Specialized Models}
\begin{figure}[h]
    \centering
    \includegraphics[width=1\linewidth]{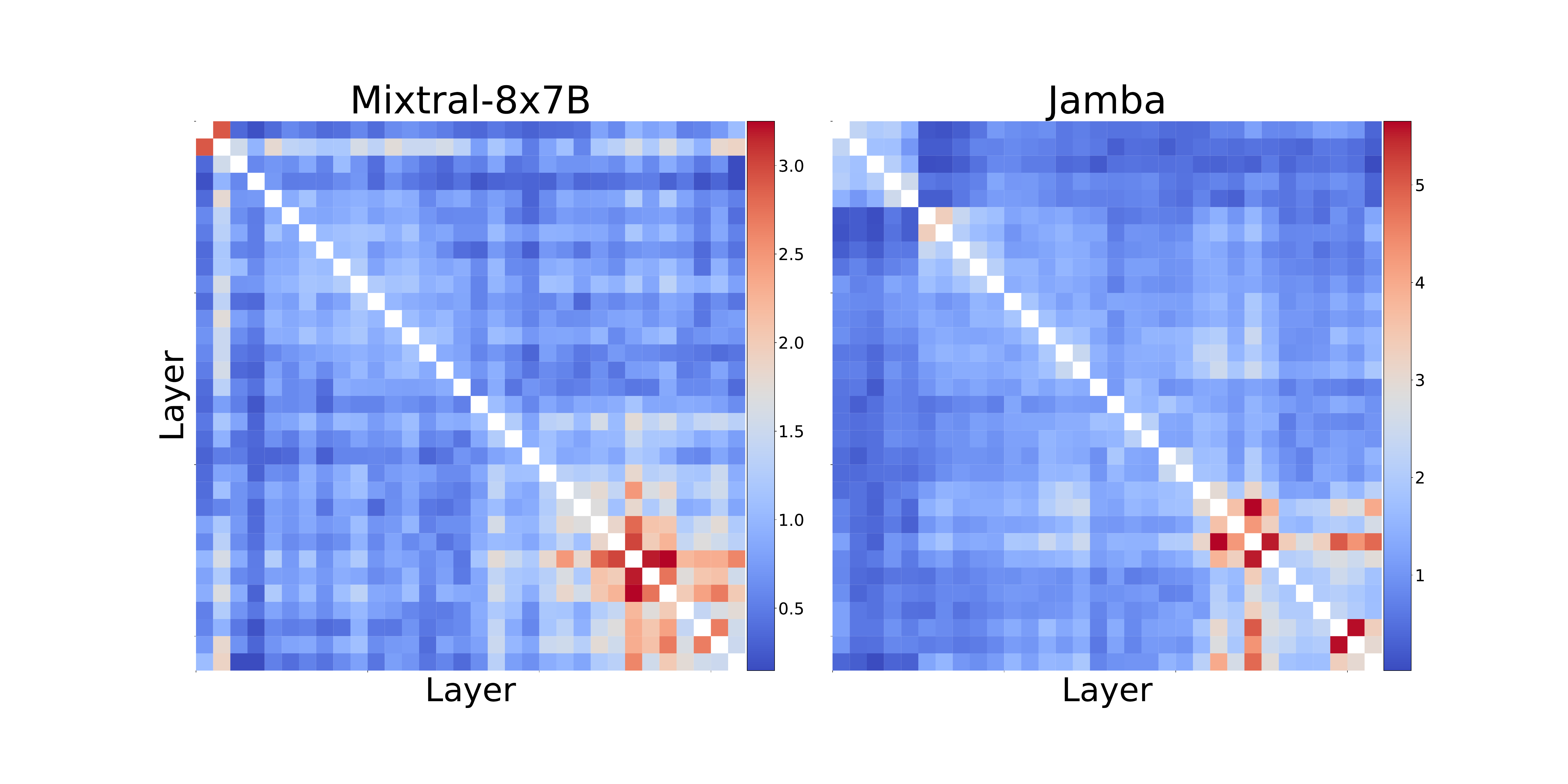}
    \caption{The similarity matrix of Mixtral-8x7B and Jamba model.}
    \label{fig:other}
\end{figure}
\begin{figure*}[!t] 
    \centering 
    \includegraphics[width=1\textwidth]{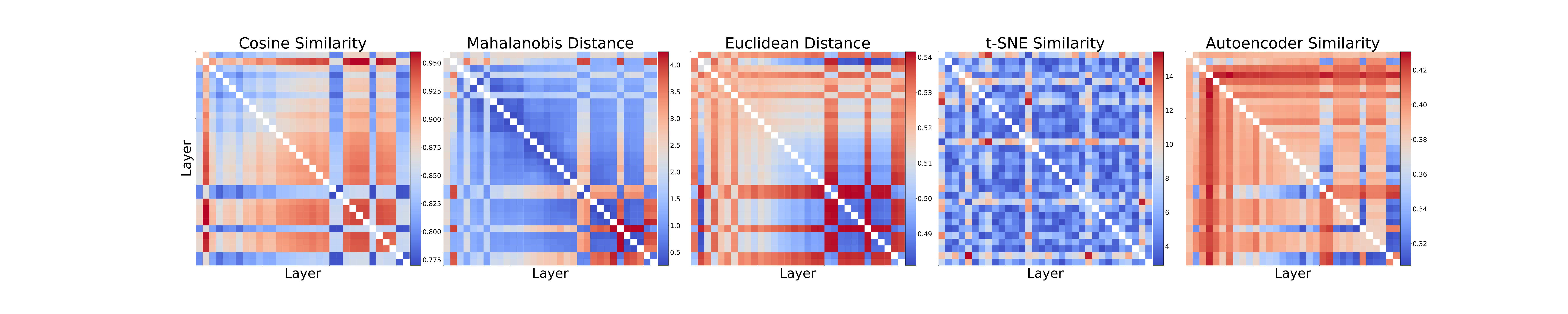}
    \caption{Similarity matrices for various measures in the Llama3-8B model, showing different patterns and effectiveness in capturing layer relationships, with none fully matching the expected merging patterns.}
    \label{fig:simother}
\end{figure*}

In addition to its application to large language models, the MKA method shows promising potential for broader adoption across a variety of deep learning architectures. This includes Mixture-of-Experts (MoE)~\cite{jiang2024mixtral}, and Jamba~\cite{lieber2024jamba} models, which can exhibit similar redundancies in their processing layers. The results show in Figure~\ref{fig:other}. Initial experiments conducted on these diverse architectures have reinforced the viability of our approach. For instance, the similarity matrices generated on Mixtral-8x7B~\cite{jiang2024mixtral} and jamba~\cite{lieber2024jamba} applying MKA have shown that our method can also be generalized to other similar models, but the similarity distributions of Mixtral-8x7B and Jamba are slightly different from LLM, and we do not yet know the reason. These experiments further validates the effectiveness of our method across different model types.

\subsection{Analysis of Similarity Measures}
In our evaluation of the Llama3-8B model, we explored several similarity measures: Cosine Similarity, Mahalanobis Distance, Euclidean Distance, t-SNE Similarity, and Autoencoder Similarity. The similarity matrices are shown in Figure~\ref{fig:simother}. From the results, we observe that Cosine Similarity, Mahalanobis Distance, and Euclidean Distance display similar distribution patterns with vertical stripes and varied heat values. However, Mahalanobis Distance shows irregular heat values within these stripes, indicating a misalignment with the fused layer data structure. t-SNE Similarity appears random and lacks consistent patterns. For Autoencoder Similarity, the high heat values do not correspond to suitable merging areas or expected high-similarity regions.

\subsection{Further Exploration of the Merging Ratio}

To further investigate the impact of $\lambda_m$, we conducted experiments using fixed values of $\lambda_m$ without considering layer similarities. We test $\lambda_m$ values of 0.7, 0.6, 0.5, and 0.4 (assigning higher weight to the lower-numbered layer). The results are shown in Table~\ref{tab:fixed_lambda}.

These results exhibit a somewhat monotonic trend, with performance decreasing as $\lambda_m$ moves away from 0.7. However, all performances remain below that of the similarity-based method. This further highlights the importance of adaptive merging ratios based on layer similarities, as in our MKA method, for maintaining model performance during compression.

\subsection{Variations in Accuracy Across Different MMLU Subjects During Layer Merging}
\begin{figure}[h]
    \centering
    \includegraphics[width=1\linewidth]{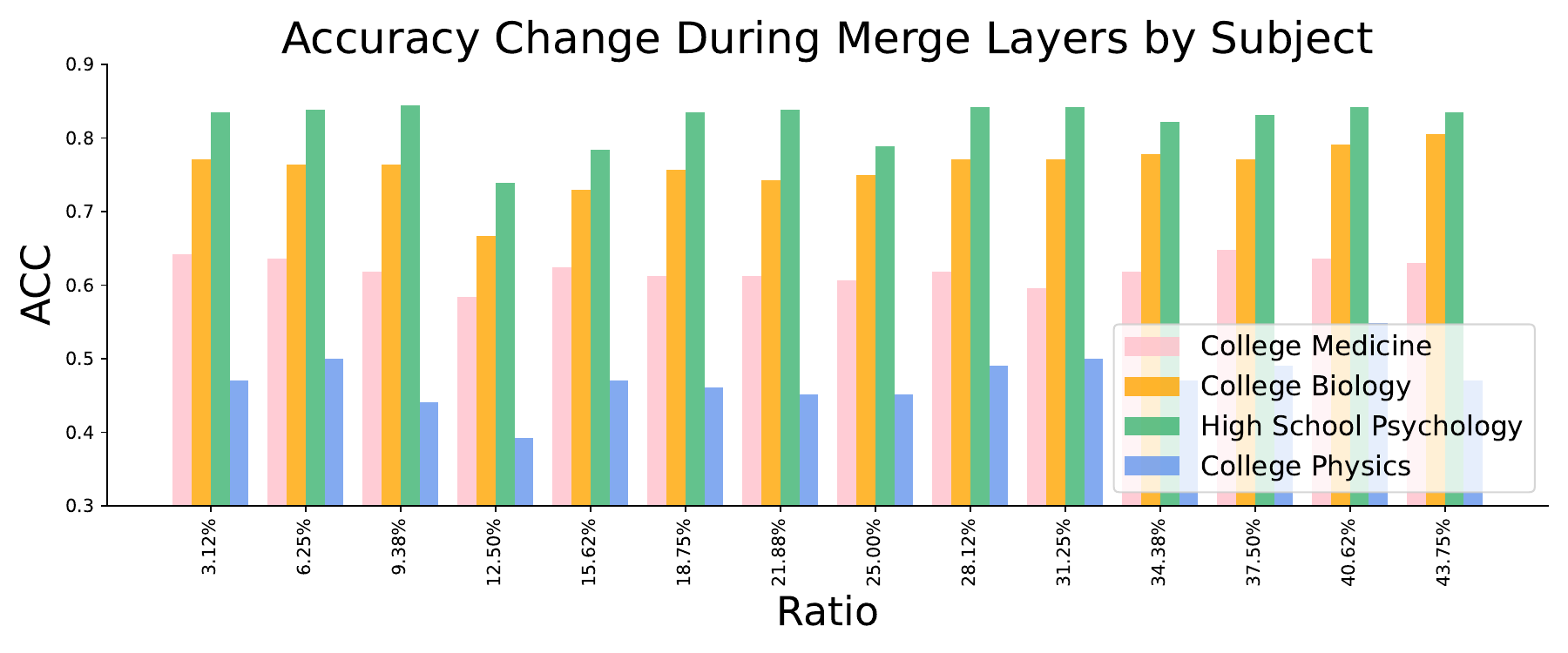}
    \caption{Different MMLU dataset subjects ACC change during merging.}
    \label{fig:subject_change}
\end{figure}
We examine the impact of model merging on performance across various academic subjects in the MMLU benchmark. Figure \ref{fig:subject_change} shows the accuracy changes across subjects such as College Medicine, College Biology, High School Psychology, and College Physics during different stages of merging model layers. From our results, we observe that High School Psychology maintained a stable accuracy with only minor fluctuations, suggesting a consistent performance and low sensitivity to the merging process. In contrast, College Biology experiences a significant drop in accuracy at the 12.5\% merging ratio, followed by a recovery. College Physics exhibits frequent fluctuations in accuracy, pointing to a high sensitivity to layer merging. Conversely, College Medicine experiences a steady increase in performance with only minor variations.

\begin{table}[h]
    \centering
    \normalsize
    \caption{Performance comparison of different fixed merging ratios on Llama3-8B using the MMLU dataset.}
    \scalebox{0.8}{
    \begin{tabular}{c|c|c|c|c|c}
        \toprule
        \textbf{CR} & MKA & $\lambda_m=0.7$ & $\lambda_m=0.6$ & $\lambda_m=0.5$ & $\lambda_m=0.4$ \\
        \midrule
        9.38 & \textbf{66.15} & 66.06(\textcolor{red}{-0.09}) & 66.05(\textcolor{red}{-0.10}) & 65.98(\textcolor{red}{-0.17}) & 65.96(\textcolor{red}{-0.19}) \\
        18.75 & \textbf{64.96} & 63.47(\textcolor{red}{-1.49}) & 63.32(\textcolor{red}{-1.64}) & 62.92(\textcolor{red}{-2.04}) & 62.83(\textcolor{red}{-2.13}) \\
        34.38 & \textbf{64.87} & 61.84(\textcolor{red}{-3.03}) & 61.52(\textcolor{red}{-3.35}) & 61.45(\textcolor{red}{-3.42}) & 61.59(\textcolor{red}{-3.28}) \\
        \bottomrule
    \end{tabular}}
    \label{tab:fixed_lambda}
\end{table}




\begin{table}[h]
\centering
\normalsize
\caption{Comparison of iterative and non-iterative MKA approaches on Llama3-8B using MMLU dataset.}
\scalebox{0.90}{
\begin{tabular}{c|c|c}
\toprule
\textbf{CR} & \textbf{MKA (w/o iterative)} & \textbf{MKA (w/ iterative)} \\
\midrule
0.00 & \textbf{66.29} & \textbf{66.29} \\
3.13 & \textbf{66.13} & \textbf{66.13} \\
6.25 & 61.64 & \textbf{66.26} \\
9.38 & 47.43 & \textbf{66.15} \\
12.50 & 35.87 & \textbf{58.08} \\
15.63 & 47.82 & \textbf{62.94} \\
18.75 & 42.01 & \textbf{64.96} \\
21.88 & 42.00 & \textbf{62.92} \\
25.00 & 39.39 & \textbf{64.28} \\
28.13 & 40.07 & \textbf{65.01} \\
31.25 & 30.41 & \textbf{63.99} \\
34.38 & 26.73 & \textbf{64.87} \\
37.50 & 25.37 & \textbf{62.05} \\
\bottomrule
\end{tabular}}
\label{tab:iterative_comparison}
\end{table}

\section{Conclusion}

In this paper, we have proposed Manifold-Based Knowledge Alignment and Layer Merging Compression (MKA), a novel model compression technique specifically designed to efficiently reduce the size of large language models (LLMs) while maintaining their performance. MKA leverages manifold learning techniques to align knowledge across layers and utilizes the Information Bottleneck (IB) measure to identify the most similar layers for merging. By capturing the intricate nonlinear dependencies within LLMs and integrating knowledge from similar layers, MKA achieves remarkable compression ratios without sacrificing model accuracy. We have conducted extensive experiments on a diverse set of benchmark datasets and various state-of-the-art LLMs to rigorously evaluate the effectiveness of MKA in preserving model performance while significantly reducing model size. Our empirical results demonstrate that MKA consistently outperforms existing pruning methods and can achieve even higher compression ratios when combined with quantization techniques.

\section*{Acknowledgements}
This work is supported by the National Key R\&D Program of China (Grant No. 2023YFB3307500).
This work is also supported by the National Natural Science Foundation of China (Grant No. 62306087 and No. 62472121), the Natural Science Foundation of
Shandong Province (Grant No. ZR2023QF154), Special Funding Program of Shandong Taishan Scholars Project and the Open Project of Anhui Provincial Key Laboratory of Multimodal Cognitive Computation, Anhui University (Grant No. MMC202420). Besides, we sincerely thank the anonymous reviewers for their valuable feedback.

\bibliography{main}
\bibliographystyle{IEEEtran}

\vfill

\end{document}